\documentclass{article}
\usepackage{spconf,amsmath,graphicx,multirow}

\title{
Soft labeling by Distilling Anatomical knowledge for Improved MS Lesion Segmentation 
}

\name{
Eytan Kats${^\star}$
\qquad Jacob Goldberger${^\dagger}$
\qquad Hayit Greenspan${^\star}$
}

\address{
${^\star}$ Department of Biomedical Engineering, Tel Aviv University, Tel Aviv, Israel \\
${^\dagger}$ Faculty of Engineering, Bar-Ilan University, Ramat-Gan, Israel 
}

\begin{document}
\maketitle

\begin{abstract}

This paper explores the use of a soft ground-truth mask (``soft mask'') to train a Fully Convolutional Neural Network (FCNN) for segmentation of Multiple Sclerosis (MS) lesions. Detection and segmentation of MS lesions is a complex task largely due to the extreme unbalanced data, with very small number of lesion pixels that can be used for training. Utilizing the anatomical knowledge that the lesion surrounding pixels may also include some lesion level information, we suggest to increase the data set of the lesion class with neighboring pixel data - with a reduced confidence weight. A soft mask is constructed by morphological dilation of the binary segmentation mask provided by a given expert, where expert-marked voxels receive label 1 and voxels of the dilated region are assigned a soft label. In the methodology proposed, the FCNN is trained using the soft mask. On the ISBI 2015 challenge dataset, this is shown to provide a better precision-recall tradeoff and to achieve a higher average Dice similarity coefficient. We also show that by using this soft mask scheme we can improve the network segmentation performance when compared to a  second independent expert.
\end{abstract}


\section{Introduction}
\label{sec:intro}

\begin{figure}[htb]
  \includegraphics[width=8.5cm]{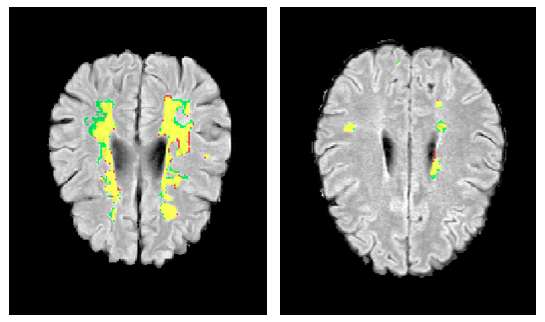}
  \caption{Examples of expert ground-truth binary masks: yellow - lesion voxels as defined by both raters, green - voxels delineated only by rater 1, red - voxels delineated only by rater 2.}
  \label{fig:raters_variability}
\end{figure}

Multiple Sclerosis (MS) is an autoimmune disease of the central nervous system that is characterized by the formation of delineated lesions visible in Magnetic Resonance Images (MRI).
Accurate segmentation of MS lesions is essential for reliable disease onset detection, in tracking its progression and in evaluating treatment efficiency. In recent years, FCNNs have achieved promising results in lesion segmentation (see e.g. \cite{3d_cnn_crf_kamnitsas}, \cite{cascaded_3d_cnn_valverde}, \cite{gru_andermatt}). A major challenge in training FCNNs for MS lesion segmentation is the highly unbalanced data, as the number of lesion class voxels are often much lower than the number of non-lesion voxels. Models trained with unbalanced data yield segmentation that is biased towards the non-lesion class which is  characterized by a greater  amount of false negatives. There are various methods to address this data imbalance, including equal sampling \cite{cascaded_3d_cnn_valverde}, 2-phase training \cite{brain_tumor_with_dnn_havaei}, and persistent loss functions \cite{generalized_dice_sudre}  \cite{tversky_loss_for_3d_cnn_sudre}. A second challenge facing delineation of MS lesions is the fact that the lesion contours are not well defined on the MRI images - leading to much ambiguity in the expert markings along the lesion contours. In Figure  \ref{fig:raters_variability} we see two MRI scans overlaid with an expert ground-truth binary mask. In order to demonstrate the ambiguity, we show masks by 2 experts: One of the experts (rater 1) labels all the yellow and green voxels as lesion voxels ("1"). A second expert (rater 2) also labels all the yellow voxels as lesion voxels, but does not find that the green voxels belong to the lesion class. We note that the conflicting and more problematic voxels lie on the boundary of the lesion area.

A loss function used for FCNN training is expected to reflect the true delineations of the expert. As most of the inter-rater variability can be found along MS lesion contour  voxels, we propose to modify the true delineations at those pixels by assigning  soft class probabilities. Specifically, we present a loss function for training the FCNN that uses soft labeling within the framework of a Dice measure.  This loss function provides additional  information for the training process beyond the ground truth mask obtained by the expert. We evaluate the proposed training loss function on the ISBI 2015 MS lesion segmentation challenge dataset \cite{longitudinal_challenge_carass}. We show that utilizing the soft labeled masks for  FCNN training leads to a better precision-recall result, with a dependence on  the volume of the soft labeled region and the soft class probability value.
  
\section{Extended Soft Labeling}
\label{sec:extended_soft_labeling}

Assume we train a FCNN for the task of lesion segmentation.  A ground truth binary labeling for each voxel is provided by a manual annotation. As noted above the labeling is extremely unbalanced since most of the voxels are labeled as non-lesion. Training with imbalanced data is very problematic especially when the training evaluation measure is classification accuracy. A well-known alternative method for  evaluating the performance of medical imaging systems is the Dice measure. However, Dice measure can also be skewed by  imbalanced data.  

Knowledge distillation is a popular method for transferring knowledge from a large model (a teacher) trained on a server to a smaller, compressed model (a student) simple enough to run on a device \cite{distilling_knowledge_hinton}. Here we applied a similar strategy to provide more information when training a lesion segmentation system. Voxels that are near contours of the segmented lesion are not well defined in MRI images, yet these voxels can carry additional anatomical information about the lesion structure. To effectively use this information during the training process we modified the manual delineations of the expert by soft labeling of the voxels in the proximity of lesion.

\subsection{Soft Labeled Mask}
\label{subsec:soft_labeled_mask}

\begin{figure}[htb]
  \centering
  \includegraphics[width=5.0cm]{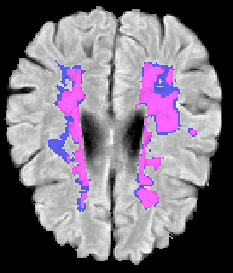}
  \caption{Example of soft mask dilated up to 140\% of the original size that we used as ground truth in FCNN training: purple - original binary mask region, blue - soft labeled voxels.}
  \label{fig:dilated_mask}
\end{figure}

To create the soft mask we expand the original binary mask by 3D morphological dilation. Using the clinical observation that lesions appear as hyper-intense regions in FLAIR images \cite{segmentation_with_spatial_consistency_mechrez}, we exclude from the dilated region those voxels with FLAIR intensity value lower than a defined threshold. Selected voxels from the dilated region are assigned a  soft label ${0<\gamma<1}$ which is interpreted as the  probability of the voxel to be part of the lesion. In the experimental section we analyze the best value for $\gamma$. The label of the manually annotated voxels remains 1. Figure \ref{fig:dilated_mask} shows an example of manual delineations of an expert and the soft labeling of pixels that are  similar to the annotated voxels in both location and content. 

\begin{figure*}[htb]
  \centering
  \includegraphics[width=15cm]{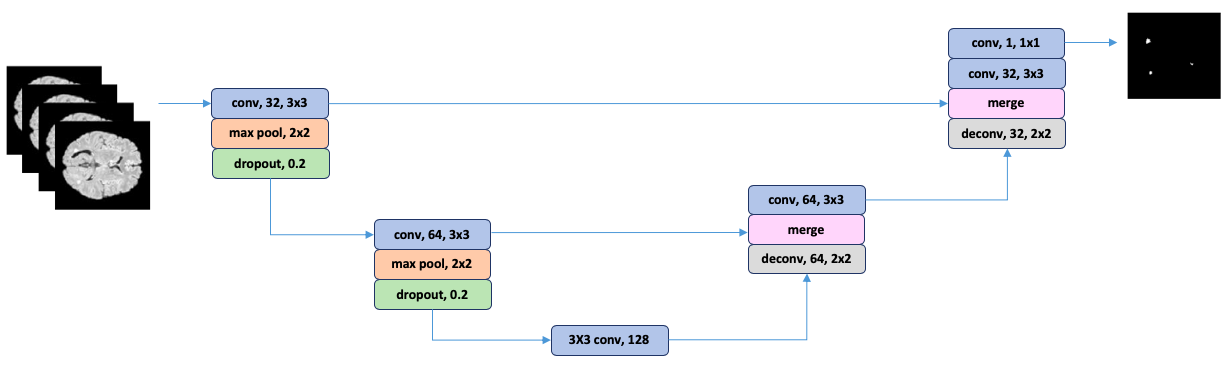}
  \caption{The 2D U-net based FCNN architecture. Each input channel is related to a different MRI modality.}
  \label{fig:network_architecture}
\end{figure*}

\subsection{Effective Loss Function}
\label{subsec:effective_loss_function}

Dice is often serves as a loss function for training  FCNN in medical imaging tasks. Dice is defined as follows:
\begin{equation}
\label{eq:dice_loss_function}
  Dice Loss = -\frac{TP}{TP + 0.5FP + 0.5FN},
\end{equation}
where TP - number of True Positive voxels, FP - number of False Positive voxels and FN - number of False Negative voxels.

By denoting ground truth mask as matrix $\textbf{T}$ and predicted probabilistic mask as matrix $\textbf{P}$ we can write
the terms in the Dice definition as follows:
$TP = \sum_{i} (\textbf{T}_{i} \cdot \textbf{P}_{i})$,
$FP = \sum_{i} \textbf{P}_{i} - \sum (\textbf{T}_{i} \cdot \textbf{P}_{i})$ and 
$FN = \sum_{i} \textbf{T}_{i} - \sum (\textbf{T}_{i} \cdot \textbf{P}_{i})$
where ${\textbf{T}_{i}}$ is a true value of the voxel $i$, and ${\textbf{P}_{i}}$ is a predicted probability of the voxel $i$.
Using  this notation the Dice loss function can be equivalently written as:
\begin{equation}
\label{eq:dice_loss_function_mask_notation}
  Dice Loss= -\frac{\sum_{i} (\textbf{T}_{i} \cdot \textbf{P}_{i})}{0.5\sum_{i} \textbf{P}_{i} + 0.5\sum_{i} \textbf{T}_{i}}.
\end{equation}

By denoting the binary mask of the dilated region as matrix \textbf{D} and the soft label assigned to voxels of the dilated region as $\gamma$ we can represent the soft labeled mask as $\textbf{T} + \gamma \textbf{D}$. As a result, training FCNN with the soft labeled mask is equivalent to training with the following effective soft dice loss function:

\begin{equation}
\label{eq:effective_loss_function}
  Soft Dice Loss= -\frac{\sum_{i} (\textbf{T}_{i} + \gamma \textbf{D}_{i}) \cdot \textbf{P}_{i}}{0.5\sum_{i} \textbf{P}_{i} + 0.5\sum_{i}(\textbf{T}_{i} + \gamma \textbf{D}_{i})}.
\end{equation}

Below we compare between training FCNN with the Dice loss (\ref{eq:dice_loss_function_mask_notation}) and the soft Dice loss (\ref{eq:effective_loss_function}) and show that when using a Dice loss based on extended soft labels for training, we obtain  better test time performance as measured by the Dice measure. 

\section{Experimental Results}
\label{sec:experimental_results}

In this section  we evaluate the proposed loss function on a MS lesion dataset. 
We first find the optimal combination of dilation size and a soft label value on the training data using a cross-validation technique (Sections \ref{subsec:dilation_size} and \ref{subsec:soft_label_value}). After finding the optimal combination of parameter values, we use this to evaluate the performance of the method on the test portion of the data (Section \ref{subsec:test_results}) .

\subsection{Dataset}
\label{subsec:dataset}

 The dataset of the ISBI 2015 MS lesion segmentation challenge \cite{longitudinal_challenge_carass} consists of 5 patients in the training set  and 14 patients in the test set. Each case has 4 modalities: T1-weighted, T2-weighted, proton density-weighted and FLAIR. All images are manually delineated by two experts. The measured inter-rater Dice overlap of 0.634 indicates high inter-rater variability (see Figure \ref{fig:raters_variability}).

To obtain the parameters that yield optimal results we evaluate the proposed method on the training set by applying a leave-one-out cross-validation approach. For each combination of parameters 5 identical models were trained on 4 patients and tested on the remaining patient in the training dataset. The final results of the 5 models were averaged to produce the final performance evaluation measures.

\subsection{Network Architecture and Training Details}
\label{subsec:network_architecture}

To demonstrate the results of the proposed approach we trained a U-net \cite{unet_ronneberger} based FCNN (see Figure \ref{fig:network_architecture}). Similar to the U-net, the network architecture we used is divided into two pathways of corresponding layers  which are connected to leverage both high- and low-level features. A contracting path alternates $3 \times 3$ convolution layers and $2 \times 2$ max-pooling layers with stride 2 for downsampling. The expansive path alternates $3 \times 3$ convolution layers and $2 \times 2$ transposed convolution layers as proposed by Brosch et al. \cite{convolutional_encoder_networks_brosch}. All convolution layers, except for the last one, are followed by a rectified linear unit (ReLU) \cite{relu_nair}. Activations of the last convolution layer are fed to a sigmoid function that produces a probabilistic segmentation map with values in the range of 0 and 1.

Manual delineations by the first rater were used as the original binary mask from which the soft labeled mask was constructed by the method described in Sec. \ref{subsec:soft_labeled_mask}. To slightly rectify the data imbalance we excluded  slices that did not contain lesions according to the ground truth mask from the training data. The network was trained on 80\% of the remaining slices, with 20\% held out for validation. During training we applied random rotations on the images where rotation angles were drawn from a uniform distribution over a  range of  ${[-5,5]}$. To produce a binary mask from the probabilistic output of the model, we applied a fixed threshold that maximized the mean dice similarity coefficient over the training set. In the post-processing step,  only connected components that contained more than ${18}$ voxels were selected to be lesions.

\subsection{Dilation Size}
\label{subsec:dilation_size}

First we evaluated our method for different dilation sizes whereas soft label value was fixed and equal to ${0.3}$. The results (see Table \ref{table:dilation_sizes}) show that the model trained with a soft mask dilated to 120\% of the manual segmentation size had the best precision-recall tradeoff and an  improved  overall Dice measure.

\begin{table}[htb]
  \centering
  \resizebox{8cm}{!}{
  \begin{tabular}{|c|c|c|c|c|c|c|} \hline
    \multirow{2}{*}{Mask size} & \multicolumn{2}{|c|}{Dice} & \multicolumn{2}{|c|}{Precision} & \multicolumn{2}{|c|}{Recall} \\ \cline{2-7}
    & Rater 1 & Rater 2 & Rater 1 & Rater 2 & Rater 1 & Rater 2\\ \hline

    100\% & 67.3 & 59.0 & 81.3 & 83.3 & 61.1 & 46.9 \\ \hline
    110\% & 67.2 & 58.4 & 81.8 & 83.3 & 58.9 & 45.8 \\ \hline
    120\% & \textbf{69.9} & \textbf{62.0} & 80.5 & 81.5 & 63.1 & 50.6 \\ \hline
    130\% & 68.5 & 61.2 & 77.5 & 80.0 & 62.9 & 50.3 \\ \hline
    140\% & 67.0 & 58.8 & 80.3 & 82.6 & 59.2 & 47.0 \\ \hline
  \end{tabular}}
  \caption{Cross-validation results on the training set for different dilation size.}
  \label{table:dilation_sizes}
\end{table}

\subsection{Soft Label}
\label{subsec:soft_label_value}

In this experiment the influence of soft label value on model performance was explored. The 120\% dilation size of the manual segmentation was used for the evaluation. Based on the results shown in  Table \ref{table:soft_labels} the soft label value $\gamma=0.3$ provided valuable information about near contour voxels during the training phase and achieved the highest Dice measure.

\begin{table}[htb]
  \centering
  \resizebox{8cm}{!}{
  \begin{tabular}{|c|c|c|c|c|c|c|} \hline
    \multirow{2}{*}{Soft label} & \multicolumn{2}{|c|}{Dice} & \multicolumn{2}{|c|}{Precision} & \multicolumn{2}{|c|}{Recall} \\ \cline{2-7}
    & Rater 1 & Rater 2 & Rater 1 & Rater 2 & Rater 1 & Rater 2\\ \hline

    0.0 & 67.3 & 59.0 & 81.3 & 83.3 & 61.1 & 46.9 \\ \hline
    0.2 & 68.3 & 59.8 & 81.4 & 83.7 & 60.5 & 47.5 \\ \hline
    0.3 & \textbf{69.9} & \textbf{62.0} & 80.5 & 81.5 & 63.1 & 50.6 \\ \hline
    0.4 & 68.9 & 61.8 & 76.7 & 80.0 & 65.2 & 52.3\\ \hline
  \end{tabular}}
  \caption{Cross-validation results on the training set for different soft label values.}
  \label{table:soft_labels}
\end{table}

\subsection{Test Results}
\label{subsec:test_results}

Next, we trained the model on the whole training set using the optimal combination of the dilated size and soft label value  equal to 120\% and 0.3 respectively. To compare  performance of suggested training strategy  (\ref{eq:effective_loss_function}) the model was also trained on the whole training set while utilizing manual annotations of the first rater as the ground truth (\ref{eq:dice_loss_function_mask_notation}). The two versions were submitted to the challenge website \footnote{https://smart-stats-tools.org/lesion-challenge}. The test results are shown in Table \ref{table:test_results}. We can see that using  the soft mask based loss function we gain a clear improvement in both Dice, precision and recall measures.  

\begin{table}[htb]
  \centering
  \resizebox{8cm}{!}{
  \begin{tabular}{|l|c|c|c|c|c|c|} \hline
    Method & Dice & Precision & Recall \\ \hline
    Binary mask & 56.0 & 82.9 & 44.6 \\ \hline
    Soft labeled mask & \textbf{57.8} & \textbf{83.8} & \textbf{46.6} \\ \hline
  \end{tabular}}
  \caption{Test set results for the optimal combination of parameters.}
  \label{table:test_results}
  \end{table}

\section{Conclusions}
\label{sec:conclusions}

In this study we proposed a new loss function that is used for training FCNN for MS lesion segmentation task where the training data is highly unbalanced. We defined a soft labeled mask that is assigned to voxels which are similar to the ground truth in both location and intensity. These voxels are usually near contour regions and tend to be labeled differently by different experts. We showed that training the FCNN with the proposed modified Dice loss function leads to  better model generalization. We demonstrate the propose method on MS lesion segmentation. This concept is general and can be harnessed to improve other medical image segmentation tasks.

\bibliographystyle{IEEEbib}
\bibliography{strings}

\end{document}